# Controlling Cloze-test Question Item Difficulty with PLM-based Surrogate Models for IRT Assessment

**Jingshen Zhang**[†]   **Jiajun Xie**[†]   **Xinying Qiu**[‡]

School of Information Science and Technology
Guangdong University of Foreign Studies, China

`{audbut0702,xie_jj03}@163.com; xy.qiu@foxmail.com`

## Abstract

Item difficulty plays a crucial role in adaptive testing. However, few works have focused on generating questions of varying difficulty levels, especially for multiple-choice (MC) cloze tests. We propose training pre-trained language models (PLMs) as surrogate models to enable item response theory (IRT) assessment, avoiding the need for human test subjects. We also propose two strategies to control the difficulty levels of both the gaps and the distractors using ranking rules to reduce invalid distractors. Experimentation on a benchmark dataset demonstrates that our proposed framework and methods can effectively control and evaluate the difficulty levels of MC cloze tests.

## 1 Introduction

Multiple-choice cloze tests are fill-in-the-blank questions that assess reading comprehension and overall language proficiency by requiring test takers to select the correct missing words from options. Table 1 gives an example test item consisting of a stem with a gap to fill, a key or answer, and three distractors.

| Stem: |
| --- |
| I knelt and put my arms around the child. Then the tears came, slowly at first , but soon she was ___ her heart out against my shoulder. |
| **Options:** |
| A. crying    B. shouting    C. drawing    D. knocking |
| **Key:** A    **Distractors:** B C D |

Table 1: A question item of MC Cloze test.

MC cloze test questions have been a focus of research because they are a common question format on standardized language proficiency exams such as TOEFL, TOEIC, IELTS, and college/high school entrance exams. In this paper, we address the research questions of generating MC cloze test of different item difficulty levels.

Prior studies on cloze test question generation have concentrated largely on distractor generation, with the goal of reproducing distractors exactly matching the benchmark datasets (Chung et al. 2020; Ren et al., 2021; Chiang et al. 2022; Wang et al. 2023). Although some studies have acknowledged the benefit of having distractors with diverse difficulty levels (Yeung et al., 2019), there has been minimal investigation into generating distractors with difficulty level different from the benchmark.

Item difficulty plays a crucial role in adaptive testing. It is a parameter that determines which questions to present to a test taker and estimates their proficiency level. Therefore, the difficulty of each item should be known beforehand so appropriate questions can be selected during the test (Susanti et al. 2017). However, only a number of works have focused on generating question items of various difficulty levels, for RC questions (Gao et al. 2019a), C-test questions (Lee et al. 2019 and 2020), and MC cloze questions (Susanti et al. 2017). This research gap is largely due to the lack of a reliable metric to evaluate the item difficulty of generated questions. Most previous study relies on human test takers and human annotation for assessing the change of difficulty levels (Susantia et al. 2017, Lee et al. 2019).

---

[†] Equal contribution

[‡] Corresponding author

| Related Research | Answer Type | Dataset | Factors to Control/Generate | | | Difficulty Control (Evaluation Method) | Difficulty Level |
|---|---|---|---|---|---|---|---|
| | | | Distractor (Selection Method) | Gap (Generation Method) | Stem | | |
| Gao et al. 2019a | R. C. | SQuAD | | | √ | Yes (RC system) | Item Level |
| Gao et al. 2019b | R. C. | RACE | √ | | | None | |
| Chung et al. 2020 | R. C. | RACE | √ | | | None | |
| Qiu et al. 2020 | R. C. | RACE | √ | | | None | |
| Felice et al. 2022 | Open Cloze | private | | √ (Electra) | | None | |
| Matsumori et al. 2023 | Open Cloze | private | | √ (gap score) | | None | |
| Lee et al. 2019 | C-test | Beiborn et al.2016 | | √ (prediction) | | Yes (Human Subject) | Item Level |
| Lee et al. 2020 | C-test | Beiborn et al.2016 | | √ (entropy) | | Yes (MLP model) | Proficiency Level |
| Susantia et al. 2017 | MC Cloze | TOEFL iBT | √ (feature-based) | | √ | Yes (Human subject) | Item Level |
| Yeung et al. 2019 | MC Cloze | Chinese sentences | √ (BERT-based ranking) | | | None | |
| Ren and Zhu, 2021 | MC Cloze | DGen | √ (featured-based L2R) | | | None | |
| Panda et al. 2022 | MC Cloze | ESL lounge | √ (BERT-based and feature-based) | | | None | |
| Chiang et al. 2022 | MC Cloze | CLOTH, DGen | √ (BERT-based and feature-based)) | | | None | |
| Wang et al. 2023 | MC Cloze | CLOTH, DGen | √ (Text2Text) | | | None | |
| **Our Research** | **MC Cloze** | **CLOTH** | **√ (BERT-based and feature-based with validity rules)** | **√ (confidence-based entropy)** | | **Yes (PLM-based IRT Assessment)** | **Item Level** |

Table 2: Recent Research on Question Generation for Language Proficiency Test

Our research has two main goals: (1) We propose strategies to generate cloze-test questions by controlling both the distractors and the gap, with consideration for reducing invalid distractors. (2) We address the problem of objective and efficient evaluation by using PLMs as subject surrogates to mimic Item Response Theory, bypassing the need for human test subjects. We will provide our dataset and codes upon request.

## 2 Related Research

The language proficiency test commonly adopts cloze tests (open or multiple-choice), C-tests, and reading comprehension (RC) to assess students' language skills. Question Generation (QG) aims to create natural and human-like questions from diverse data sources. Research on MC cloze test question generation primarily focuses on tasks such as analyzing factors influencing item difficulty (Susanti et al., 2017), distractor generation (Yeung et al., 2019; Ren and Zhu, 2021; Chiang et al., 2022), and reducing invalid distractors (Zesch and Melamud, 2014; Wojatzki et al., 2016). Table 2 presents a comparative analysis of recent studies on the automatic generation of cloze test, RC, and C-test.

For MC cloze test, distractor generation algorithms aim to identify plausible but incorrect candidates for filling in blanks. Selection is based on semantic proximity to the target word, measured through methods like WordNet (Brown et al., 2005), thesauri (Smith et al., 2010), and word embeddings similarity (Guo et al., 2016; Susanti et al., 2015; Jiang and Lee, 2017). Recent studies utilize confidence scores from BERT models (Devlin et al. 2018) for ranking distractor candidates, outperforming semantic similarity methods in correlation with human judgment (Yeung et al., 2019). Ren and Zhu (2021) apply knowledge-based techniques to help generate distractor candidates. Chiang et al. (2022) suggest BERT-based methods as superior in distractor generation. Their candidate selection relies on confidence scores from pretrained language models. Wang et al. (2023) propose a Text2Text formulation using pseudo Kullback-Leibler divergence, candidate augmentation and multi-task training, enhancing performance in generating distractors that align with benchmarks.

Item difficulty is crucial in adaptive testing, yet few studies focus on generating items with diverse difficulty levels different from standard benchmark datasets. Furthermore, these works typically rely on human test-taker evaluations (Susanti et al., 2017; Lee et al., 2019). A few studies used model judgments in RC test (Gao et al., 2019) and C-test (Lee et al., 2020). In related research on question difficulty estimation, QA models are also proposed to estimate difficulty through item response theory (Benedetto, 2022).

Gap generation has been the focus in the context of C-tests (Lee et al. 2019 and 2020). In open cloze tests, Felice et al. (2022) recommend transformer models and multi-objective learning for gap prediction. Matsumori et al. (2023) propose a masked language model approach with a gap score metric for generating open cloze questions tailored to specific target words. In contrast, research addressing the control of difficulty levels by modifying both distractors and gaps in multiple-choice cloze tests is lacking.

## 3 Methodology

Our research tackles the key challenges as reviewed above in the generation of MC cloze test questions. Firstly, we aim to produce questions with varying difficulty levels by effectively managing both the gap and the distractors. This involves implementing ranking rules to eliminate invalid distractors. Secondly, we propose a PLM-based IRT assessment framework for objectively evaluating difficulty changes at the item level. This evaluation method alleviates the reliance on human subjects and annotations, different from the predominant approach in previous studies.

As shown in Figure 1, our research structure contains three main components: (1) We train various PLM-based models using the original benchmark dataset to simulate human test-takers' performances; (2) We design difficulty control strategies involving manipulating the gap and distractor selection for the test data in order to generate items of varying difficulty levels; (3) We use the same PLM-based surrogate models to take the modified-level tests, and implement an IRT model to evaluate and compare the change in item difficulty before and after applying the control strategies.

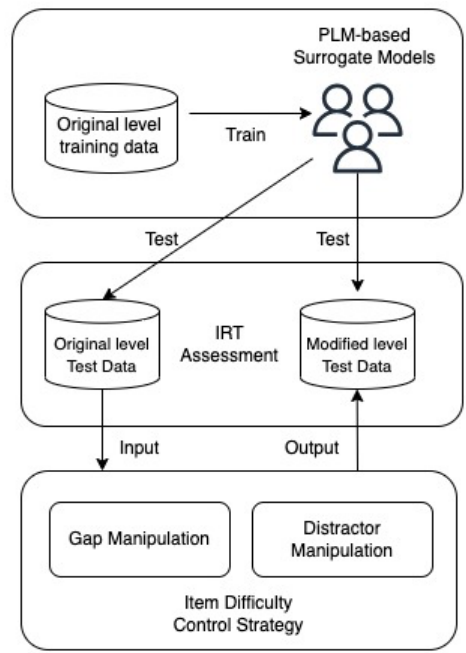

Figure 1: Research Structure

### 3.1 IRT Assessment with PLM-based Surrogate Models

Traditionally, calibrating test difficulty requires extensive trials with real human subjects. This process is time-consuming, costly, and does not easily scale. Item response theory (IRT) is a psychometric framework to estimate the difficulty of test items in an unsupervised manner (Benedetto, 2022; Susanti et al., 2017).

Similar to previous study using trained models to evaluate change of difficulty (Gao et al. 2019a, Lee et al. 2020), we suggest that the predictions by various PLMs with different parameter settings can simulate human test taking, providing informative signals for IRT without needing to recruit actual test takers.

We fine-tune 12 PLM models on the training splits of each dataset based on BigBird (Zaheer et al. 2020) and Electra (Clark et al. 2020) architectures and with different hyperparameter settings. We apply the control strategies to generate different versions for each fold of the test data. We have each trained surrogate model take the hard or easy version of their corresponding test fold and aggregate the scores across folds to obtain overall performance on the full test data.

An IRT model is fitted on the surrogate models' aggregated scores on the original test and the modified hard/easy test. By modeling the score distributions, we can evaluate shifts in difficulty levels between the easy and hard versions.

## 3.2 Difficulty-controllable Question generation

For difficulty-controllable question generation, we combine PLM-based confidence scores, semantic similarity and edit distance metrics, and validity rules to generate gaps and distractors at tunable difficulty levels.

**Gap Difficulty Control**:

Entropy has been studied as a proxy for gap complexity in open cloze tests (Felice et al., 2022) and C-tests (Lee et al., 2020). We propose to leverage pre-trained model's confidence scores for entropy estimation of a candidate gap, without separate training. The algorithm is given in Appendix A

Specifically, given a cloze question stem, we identify candidate gap words in the stem matching the part-of-speech tag of the original key that fills the stem gap. We finetune a model such as BERT to predict words for each candidate gap.

For each candidate gap, we take the top K predictions ordered by the model's confidence score. Using the top $K$ words and their scores, we calculate the Shannon entropy of the candidate gap as:

$$H(X) = -\sum_{i=1}^{n} p(x_i) \log_2 p(x_i)$$

where $x_i$ is the $i_{th}$ word predicted by BERT for the candidate gap, and $p(x_i)$ is BERT model's confidence score for $x_i$.

We sort candidate gaps by decreasing entropy and select high entropy ones for hard questions and low entropy for easy questions. Hard questions are generated by selecting hard gaps and generating more difficult distractors for the selected gaps, and vice versa for easy questions.

**Distractor Difficulty Factors**:

Distractor generation algorithms aim to identify plausible but incorrect candidates for filling in blanks in sentences. Inspired by previous studies (Susantia et al. 2017, Yeung et al. 2019, Ren and Zhu 2021, Chiang et al. 2022), we design three control factors - semantic similarity based on the cosine similarity of word2vec embeddings, syntactic similarity based on Levenshtein distance, and confidence scores from pretrained language models (PLMs) predicting the gap.

- **Confidence score**

Formally, let $\mathbb{M}()$ be a PLM model finetuned with our training data set, $S$ be a cloze stem, $V$ be a vocabulary list, $A$ be the answer of $S$, and $d_i$ be a word in $V$ as a candidate distractor. We denote a given stem $S$ with the cloze blank filled in $[Mask]$ with $S_{\otimes[Mask]}$.

Confidence score $C_i$ for $d_i$ given by PLM is defined:

$$C_i = p(d_i | \mathbb{M}(S_{\otimes[Mask]}))$$

- **Semantic similarity**

The semantic similarity $S_i$ of the candidate distractor and the answer is defined as:

$$S_i = CosineSimilarity(Embed(d_i), Embed(A))$$

where $Embed()$ refers to Glove Embedding.

- **Levenshtein ratio**

The Levenshtein ratio measures string similarity on a scale from 0 to 1. It is defined as:

$$Levenshtein_{ratio} = \frac{sum - ldist}{sum}$$

where *sum* is the total length of two strings, and *ldist* is the weighted edit distance between two strings based on Levenshtein distance (Levenshtein et al. 1966). The Levenshtein distance counts insertions, deletions, and substitutions to transform one string into the other. The weighted distance is calculated as:

$$ldist = Num(INSERT) + Num(DELETE) + 2 * Num(REPLACE)$$

A higher Levenshtein ratio indicates more similarity between the strings, with 1 being identical strings and 0 being completely different. The metric is useful for fuzzy string comparison and matching.

**Invalid Distractor Control**

Challenging distractors typically exhibit higher semantic similarity to correct answers. However, selection strategies based solely on semantic scores or pretrained language model (PLM) confidence values may generate invalid distractors - words that could also plausibly fill the gap.

As suggested in previous work (Zesch and Melamud, 2014), context-sensitive lexical inference rules can help filter distractors that are potentially appropriate options for the gap context. Our analysis of PLM-predicted distractor validity on a gap-fill dataset reveals issues. Fine-tuning

BERT and generating distractors by ranking confidence scores show 302 items with 482 invalid distractors. Among them, 365 invalid distractors were ranked higher than the correct answer by BERT in terms of confidence score.

Motivated by these observations and previous study on choosing valid distractors by considering lists of context-(in)sensitive candidates (Zesch and Melamud, 2014), we design distractor selection validity rules as follows:

(1) Valid distractors should have lower confidence ranking than answers.

(2) For more difficult items, top two distractors by semantic similarity and highest Levenshtein ratio are selected from the 50 ranks after the answer per BERT.

(3) For easier items, bottom two by semantic similarity and lowest Levenshtein ratio are chosen from ranks 50-100 after the answer.

Details about annotation and validity rule impact analysis are in Appendix B.

**Distractor Selection Strategy**

With the three defined control factors and validity rules, we design two strategies for generating challenging or easy distractors. For distractor generation, we use BERT to rank and score all candidate words, then select the top 100 ranks after the correct answer to form the distractor candidate list, implementing the first validity rule.

We have two distractor selection strategies. The first, outlined in Algorithm 2 in Appendix A, simply chooses the top 3 highest-confidence distractors from the candidate list to generate difficult questions, and the bottom 3 for easier questions. We term this the **Confidence-Ranking Control**.

The second selection strategy combines all three control factors – confidence scores, word2vec embedding similarity, and Levenshtein distance – along with the associated validity rules 2 and 3, as detailed in the Algorithm 3 in Appendix A. This integrated approach allows us to tune distractor difficulty. We term this the **3-Factor Ranking Control**.

## 4 Experiment Design

This section presents the experimentation details for validating our proposed framework and methods. Table 6 provides generation examples referencing the original item shown in Table 1.

### 4.1 Dataset

A variety of datasets have been used for cloze test generation (Table 1). Recently, CLOTH (Xie et al., 2017) and DGen (Ren & Zhu, 2021) have become popular choices. DGen compiles science questions of wide range of subjects from diverse sources spanning elementary to college level. CLOTH comprises two sets of cloze-style English reading comprehension questions authored by teachers for middle-school and high-school entrance exams. As our research aims to control item difficulty for adaptive testing, we selected the CLOTH benchmark dataset closely aligned with our study goals. However, we believe our framework and control strategy could generalize to adaptive testing in other subjects.

We divided the CLOTH dataset into two sets according to its two proficiency levels – CLOTH-M for middle school and CLOTH-H for high school entrance exams. Each set was further segmented into 5 folds. Within each fold, we split the passages into stems. Stems comprised consecutive sentences leading up to the first [MASK] token (i.e. gap), ensuring sufficient context surrounding the cloze deletion. Table 3 presents the number of items per split in our dataset.

| Fold | Split | CLOTH-M | CLOTH-H |
|---|---|---|---|
| 0 | Train | 17123 | 42540 |
| | Validate | 5678 | 14189 |
| | Test | 5669 | 14139 |
| 1 | Train | 16975 | 42432 |
| | Validate | 5757 | 14155 |
| | Test | 5738 | 14281 |
| 2 | Train | 17011 | 42628 |
| | Validate | 5680 | 14145 |
| | Test | 5779 | 14095 |
| 3 | Train | 17094 | 42502 |
| | Validate | 5733 | 14194 |
| | Test | 5643 | 14172 |
| 4 | Train | 17077 | 42463 |
| | Validate | 5752 | 14224 |
| | Test | 5641 | 14181 |

Table 3: Data Statistics

### 4.2 Evaluation

For each data fold, we trained 12 PLM models using BigBird and Electra architectures, with learning rates of 1e-4, 1e-5, and 3e-5, batch sizes of 16 and 32, epoch of 1 and AdamW optimizer. We conducted experiments on a single NVIDIA Quadro RTX 8000 GPU. The control strategies were applied to the "Test" split. By concatenating the scores across all surrogate models and test folds, IRT models were then fitted to quantify overall changes in test difficulty. We use the py-irt library (Lalor and Rodriguez, 2023) as it leverages PyTorch and GPU acceleration for faster and more scalable IRT modeling compared to existing libraries. We apply the 1PL (also known as the Rash model) with default setting. This model estimates a latent ability parameter for subjects and a latent difficulty parameter for items, which fits exactly what we intend to evaluate.

## 5 Results and Analysis

**Surrogate Model Performance**

We first present the surrogate models' average accuracies across the five folds of test data on the original cloze items, as shown in Table 3. The yellow highlights indicate the performances of 12 surrogate test takers for CLOTH-M, while the blue highlights indicate the performances of another 12 surrogates for CLOTH-H. We see that the surrogate models have a wide range of accuracies on the original tests, spanning from 0.42 to 0.81. This demonstrates that the models have diverse capabilities to serve as artificial test takers for difficulty modeling.

| Proficiency | CLOTH-M | | CLOTH-H | |
|---|---|---|---|---|
| Model | BigBird | Electra | BigBird | Electra |
| 1e-4, 16 | 0.4282 | 0.7106 | 0.4234 | 0.527 |
| 1e-4, 32 | 0.6691 | 0.7306 | 0.5671 | 0.6601 |
| 1e-5, 16 | 0.811 | 0.7613 | 0.7902 | 0.7119 |
| 1e-5, 32 | 0.8081 | 0.7602 | 0.7974 | 0.7102 |
| 3e-5, 16 | 0.6093 | 0.7558 | 0.687 | 0.7008 |
| 3e-5, 32 | 0.798 | 0.7615 | 0.7814 | 0.7072 |

Table 4. Surrogate models' performance on the original tests.

To further study the performances of these surrogate models, we select 4 of them as surrogates for middle school test takers (i.e. Electra (1e-4, 16), Electra (1e-4, 32), Bigbird (1e-4, 32), and Electra (3e-5, 32)), and 3 of them as surrogates for high school test takers (i.e. Bigbird (1e-5, 16), Bigbird (1e-5, 32), Bigbird (3e-5, 32)). We train the 4 middle-school models and the 3 high-school models similarly and have them take both the CLOTH-M and the CLOTH-H tests. The following table compares the average accuracies, standard deviation, and utility ratios of the two sets of 12 surrogates, the middle-school surrogates and the high-school surrogates:

| Model | CLOTH-M | | | CLOTH-H | | |
|---|---|---|---|---|---|---|
| | Avg. Acc. | Stdv | Utility Ratio | Avg. Acc. | Stdv | Utility Ratio |
| 12 | 0.717 | 0.104 | 73.9% | 0.672 | 0.108 | 72.1% |
| 4-mid | 0.718 | 0.034 | 38.2% | 0.615 | 0.072 | 52.2% |
| 3-high | 0.803 | 0.006 | 10.4% | 0.79 | 0.007 | 10.9% |

Table 5. Comparing surrogate models

Utility ratio is the percentage of test questions remaining after excluding those answered correctly or incorrectly by all test takers. Table 4 shows the 4 middle school surrogates perform better on CLOTH-M and worse on CLOTH-H, while the 3 high school surrogates substantially outscore them on CLOTH-M. The smaller standard deviations demonstrate these sets represent distinct proficiency levels. However, the 12-surrogate sets achieve higher utility ratios (73.9%, 72.1%) than the middle and high school sets. Therefore, the 12-surrogate sets are retained for evaluating item difficulty control given their better utility and diverse performances to distinguish between stronger and weaker students.

**Performance of Control Methods on Two Proficiency-level Datasets**

We demonstrate the effect of generating difficult and easy items using the Confidence-ranking algorithm and 3-Factor strategy. Figures 2 present results for CLOTH-M, Figures 3 for CLOTH-H. Red lines show IRT distributions for difficult generated items, blue for easy, black dotted lines mark the original test difficulty.

Both strategies systematically manipulate cloze item difficulty. Across CLOTH-M and CLOTH-H, the strategies successfully generate harder items (red distribution shift right) and easier items (blue shift left) versus the original test items.

However, CLOTH-H exhibits a narrower spread between high/low difficulty items. This indicates greater efficacy adjusting difficulty for the lower proficiency CLOTH-M rather than the advanced CLOTH-H items.

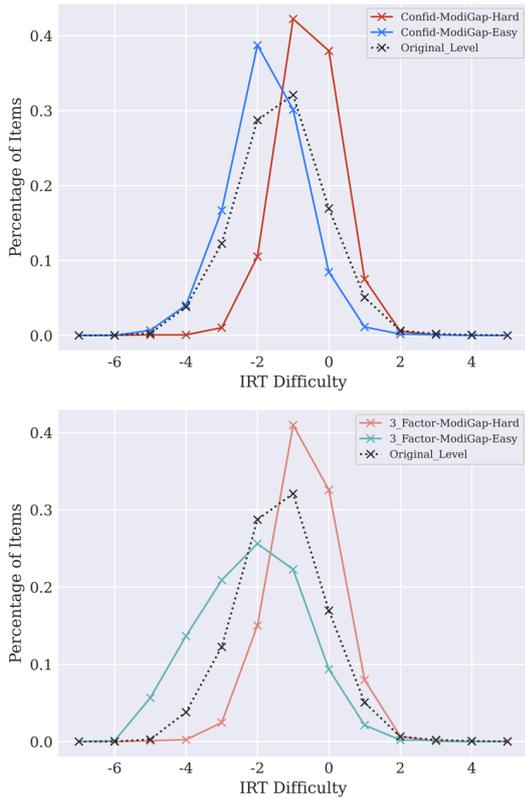

Figure 2. Change of IRT for CLOTH-M with Confidence-Ranking Control (above) and 3-Factor Ranking Control (below)

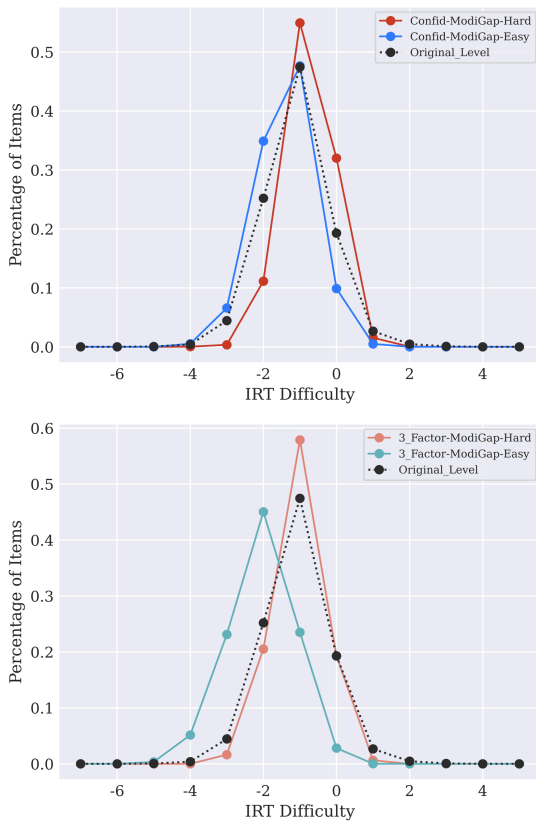

Figure 3. Change of IRT for CLOTH-H with Confidence-Ranking Control (above) and 3-Factor Ranking Control (below)

**Effect of Gap Control**

The effect of gap position control on difficulty control differs for intermediate (CLOTH-M) versus advanced (CLOTH-H) questions. As shown in Figures 4 and 5 for CLOTH-M, retaining the original gap position versus modifying it does not substantially impact the efficacy of confidence ranking (Fig. 5) or 3-factor ranking (Fig. 4). The item difficulty distributions remain relatively consistent.

In contrast, for CLOTH-H, retaining the original gap positions leads to wider item difficulty distributions for both strategies when generating hard questions, as shown with the rose dashed line in Figure 6 and the red dashed line in Figure 7. However, for generating easy items for CLOTH-H, controlling the gap position or not generates similar difficulty distributions.

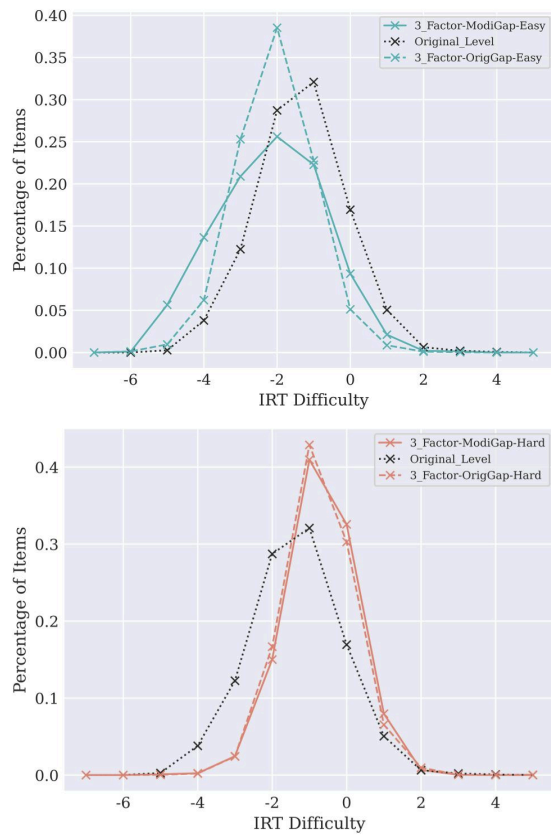

Figure 4. Effect of 3-factor Ranking on CLOTH-M with (solid line) and without (dashed line) Gap-control for easy question (green above) and hard question (rose below) generation.

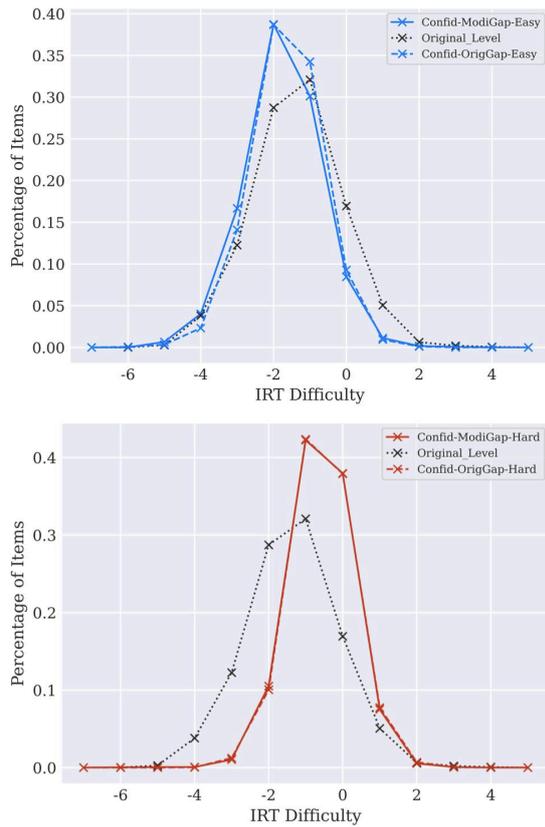

Figure 5. Effect of Confidence Ranking on CLOTH-M with (solid line) and without (dashed line) Gap-control for easy question (blue above) and hard question (red below) generation.

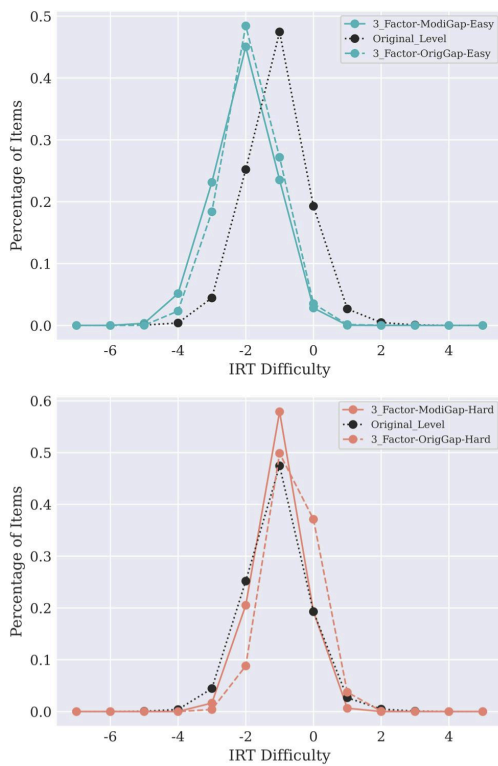

Figure 6. Effect of 3-factor Ranking on CLOTH-H with (solid line) and without (dashed line) Gap control for easy question (green above) and hard question (rose below) generation.

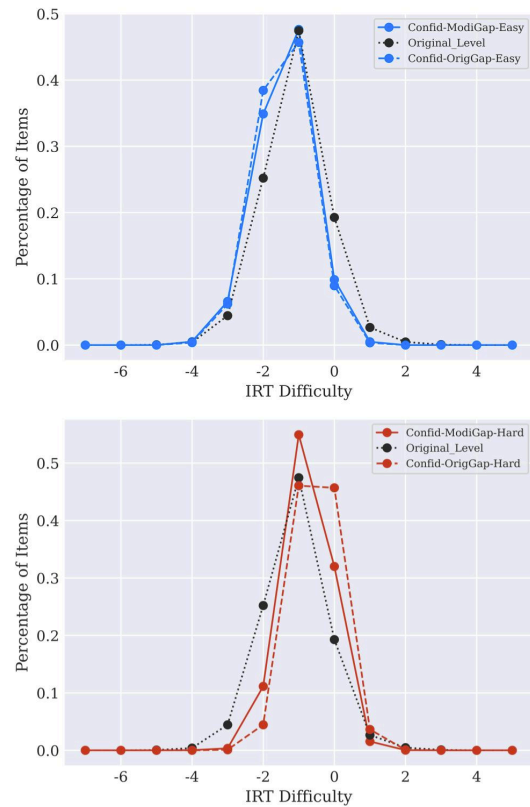

Figure 7. Effect of Confidence Ranking on CLOTH-H with (solid line) and without (dashed line) Gap control for easy question (blue above) and hard question (red below) generation.

## Comparing Confidence-ranking and 3-Factor Ranking on Generating Easy and Hard Items

Comparing the two control strategies, 3-factor ranking generates a slightly wider range of easy item difficulties for both datasets as shown in Figure 8. Meanwhile, confidence-ranking method without gap control produces slightly wider distributions of hard items for both datasets as evident in Figure 9.

## Best Combination Strategies and Advantage of Gap Control

We provide the box plot analysis on best combination control strategies for the two proficiency tests. Figures 10 and 11 show that the best strategy combination is the 3-Factor Ranking control without gap for easy question and confidence ranking control without gap for hard questions. Figure 12 shows Gap Control with 3-Factor ranking enhances easy CLOTH-M item generation over 3-Factor without gap control. While maintaining the same mean difficulty, Gap Control increases variability, indicating improved ability to span multiple difficulty values - better fulfilling key needs when creating easy test questions.

|  | Distractor generation w/ Confidence-Ranking Control | Distractor generation w/ 3-Factor Ranking Control |
|---|---|---|
| Hard | (I) **Stem:** I knelt and put my arms around the child. Then the tears came, slowly at first , but soon she was ___ her heart out against my shoulder. **Options:** A. crying  B. sobbing  C. pouring  D. weeping **Key:** A    **Distractors:** B C D | (II) **Stem:** I knelt and put my arms around the child. Then the tears came, slowly at first , but soon she was ___ her heart out against my shoulder. **Options:** A. crying  B. screaming  C. cried  D. crushed **Key:** A    **Distractors:** B C D |
| Easy | (III) **Stem:** I knelt and put my arms around the child. Then the tears came, slowly at first , but soon she was ___ her heart out against my shoulder. **Options:** A. crying  B. counting  C. shouting  D. booming **Key:** A    **Distractors:** B C D | (IV) **Stem:** I knelt and put my arms around the child. Then the tears came, slowly at first , but soon she was ___ her heart out against my shoulder. **Options:** A. crying  B. owing  C. caves  D. sobbed **Key:** A    **Distractors:** B C D |

Table 6: Generated hard and easy items for the original item in Table 1

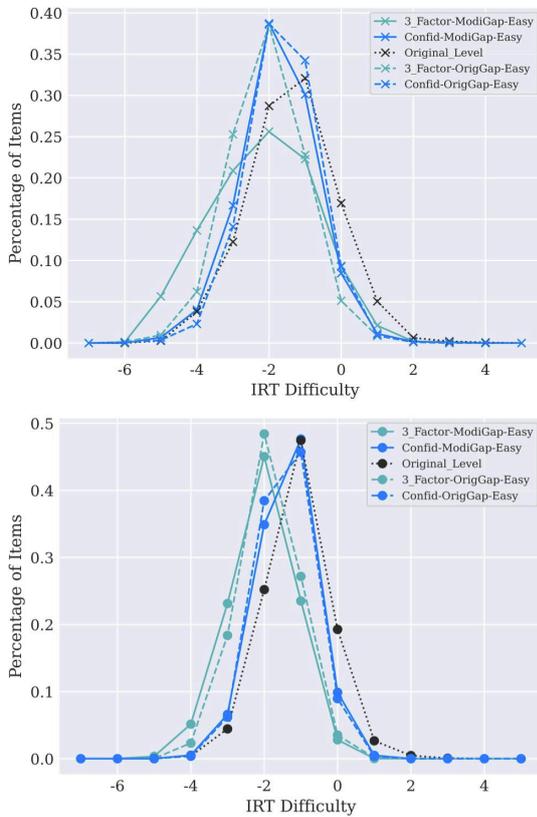

Figure 8: Confidence-ranking and 3-factor ranking w/ and w/o Gap control on generating easy items for CLOTH-M (above) and CLOTH-H (below)

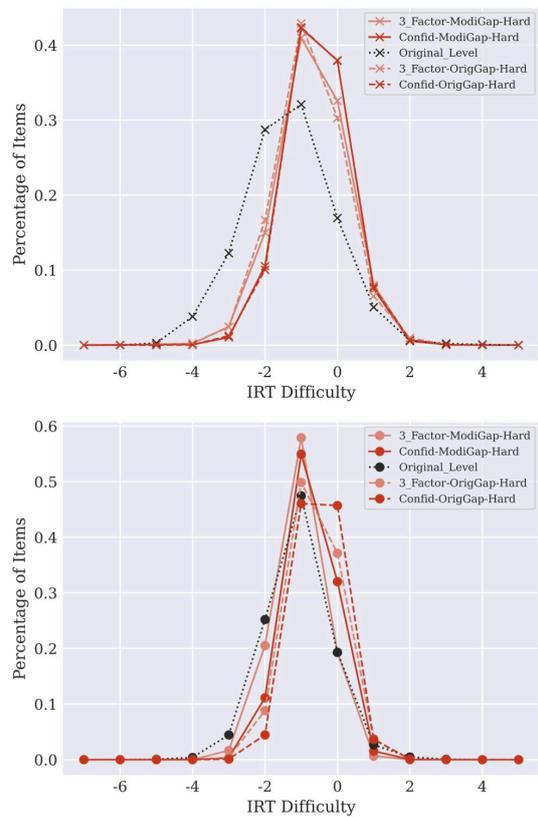

Figure 9: Confidence-Ranking and 3-Factor Ranking w/ and w/o Gap control on generating hard items for CLOTH-M (above) and CLOTH-H (below)

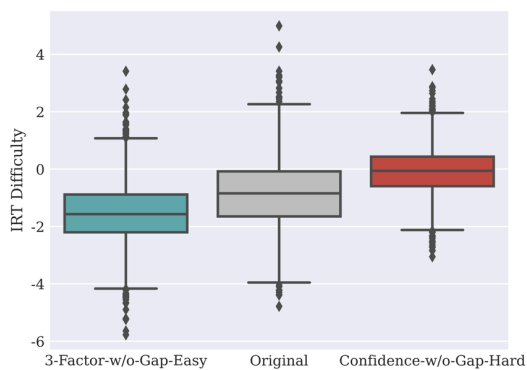

Figure 10: Best combination for CLOTH-M: 3-Factor Ranking without Gap Control for Easy Items and Confidence-Ranking without Gap Control for Hard Items Generation.

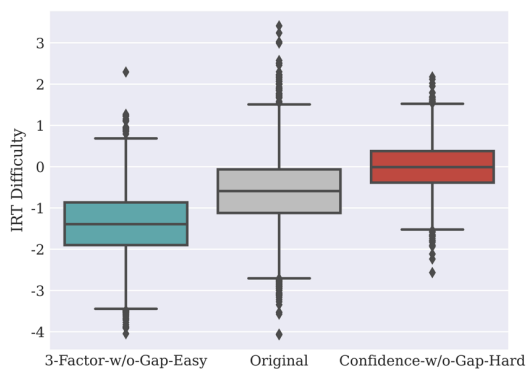

Figure 11: Best combination for CLOTH-H: 3-Factor Ranking without Gap Control for Easy Items and Confidence-Ranking without Gap Control for Hard Items Generation.

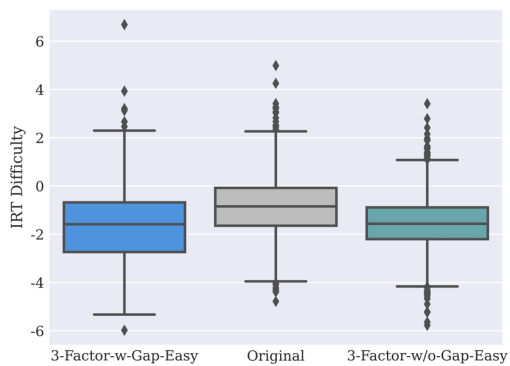

Figure 12: Gap Control strategy increases item variability than without gap control for 3-Factor Ranking method on easy item generation for CLOTH-M.

## 6 Conclusions

In this work, we proposed a novel evaluation framework for assessing control of item-level difficulty for MC Cloze test. By using diverse pretrained models as surrogate test takers, we fitted IRT distributions to quantify changes in difficulty - avoiding reliance on human test subjects.

We designed two strategies leveraging entropy, semantic similarity, edit distance to manipulate both the gap position and distractor selection for difficultuy-controlled question generation. We further implemented validity rules to reduce generation of invalid distractors.

Systematic experimentation shows: (1) The advanced test (CLOTH-H) is more difficult to control than intermediate test (CLOTH-M); (2) Gap control has a limited effect, yet increases item variability for easy CLOTH-M generation; (3) Comparatively, 3-Factor Ranking Control method works better for easy items generation while Confidence Ranking Control method exceeds at hard item generation; (4) Validity rules reduce but do not eliminate invalid distractors -- further study into this challenge is desired.

## 7 Limitation

Our difficulty control methods worked better for intermediate exam questions than advanced ones. More research is needed to improve the methods' ability to handle very complex test items. Additionally, our techniques should be validated across other subject domains. Questions also persist around optimizing validity methods to avoid invalid distractors.

## A  Algorithms

Figures 13, 14, and 15 present the algorithms for Gap Control, Confidence-Ranking Control, and 3-Factor Ranking Control respectively.

```
Algorithm 1: Gap generation
  Input  : A sentence list with POS A = [s_1, ..., s_n], Dict of POS
           numbers for answer words D_POS, Prediction model M,
           Numbers of model candidate K, Target level L
  Output: Target sentence list T
1 T ← [];
2 for key in D_POS do
3  │ AllSentenceDict[key] ← [];
4 end
5 for sentence in A do
6  │ for word in sentence do
7  │  │ pos_word = pos(word);
8  │  │ if pos_word in D_POS then
9  │  │  │ sentence_mask = CreateMaskSentence(sentence, word);
10 │  │  │ PredList = ModelPred(sentence_mask, K, M) ;  // predict
                the top K score of sentence_mask Using M
11 │  │  │ shannon = CalShannon(PredList) ;  // compute Shannon
                entropy of sentence_mask
12 │  │  │ AllSentenceDict[pos_word].append([sentence_mask, shannon])
13 │  │ end
14 │ end
15 end
16 for key in AllSentenceDict do
17 │ value = D_POS[key];
18 │ if L is Hard then
19 │  │ T ← Top value candidate sentences sorted by shannon in
              allSentenceDict[key]
20 │ else if L is Easy then
21 │  │ T ← Last value candidate sentences sorted by shannon in
              allSentenceDict[key]
22 │ end
23 end
```

Figure 13: Gap Generation Algorithm

```
Algorithm 2: Distractor Generation with Confidence Ranking
  Input  : A sentence S with a cloze, Answer Word A, Prediction Model
           M, Vocabulary List V, Numbers of Candidate K
  Output: Difficult distractors D_hard, Easy distractors D_easy
1 D_hard, D_easy ← {}, {};
2 ConfidenceList ← sorted(ModelPred(S, M, A, V)) ;  // Sort
    vocabulary V based on the scores predicted by the model M
3 Position_A ← ConfidenceList.index(A);
4 CandidateList = ConfidenceList[Position_A + 1 : Position_A + K];
5 D_hard ← Top 3 candidates by CandidateList;
6 D_easy ← Last 3 candidates by CandidateList;
7 return D_hard, D_easy
```

Figure 14: Distractor Generation with Confidence-Ranking Control

```
Algorithm 3: Distractor Generation with 3-Factor Ranking
  Input  : A sentence S with a cloze, Answer Word A, Prediction Model
           M, Vocabulary List V, Numbers of Candidate K
  Output: Difficult distractors D_hard, Easy distractors D_easy
1 D_hard, D_easy ← {}, {};
2 ConfidenceList ← sorted(ModelPred(S, M, A, V)) ;  // Sort
    vocabulary V based on the scores predicted by the model M
3 Position_A ← ConfidenceList.index(A);
4 CandidateList ← ConfidenceList[Position_A + 1 : Position_A + K];
5 Similarity_Glove, Similarity_Leven ← [], [];
6 for word in CandidateList do
7  │ G ← Calculate Glove similarity(A, word);
8  │ L ← Calculate Leven similarity(A, word);
9  │ Similarity_Glove.append(G);
10 │ Similarity_Leven.append(L);
11 end
12 GloveSimList_hard, GloveSimList_easy ←Similarity_Glove[Position_A + 1 :
    Position_A + K/2], Similarity_Glove[Position_A + K/2 : Position_A + K];
13 LevenSimList_hard, LevenSimList_easy ←Similarity_Leven[Position_A + 1 :
    Position_A + K/2], Similarity_Leven[Position_A + K/2 : Position_A + K];
14 D_hard ← Top 2 candidates by GloveSimList_hard, Top 1 candidate by
    LevenSimList_hard;
15 D_easy ← Last 2 candidates by GloveSimList_easy, Last 1 candidate by
    LevenSimList_easy;
16 return D_hard, D_easy
```

Figure 15: Distractor Generation with 3-Factor Ranking Control

## B  Annotation for Invalid Distractor Control

We analyzed the issues of invalid distractors with human evaluation. We recruited 9 college students at the CET-6 English proficiency level as annotators. The invalid distractors will most likely appear when generating hard items. Using BERT's confidence score ranking without validity control, we generated distractors for 4,575 items randomly selected from the CLOTH-H dataset. Manual annotation identified 1,676 items as having at least one invalid generated distractor (i.e., a distractor that could fit as an answer in the gap). As our control strategies involves ranking distractors after the answer, we identified 302 items to further test validity rules. Among the 906 distractors generated, 482 were annotated as invalid, representing an invalidity ratio of 53.2%. After applying the Confidence-Ranking Control method and 3-Factor Ranking Control method, the ratios dropped to 20.3% and 17.3% respectively (Table 7).

| Strategy | Num. of Invalid Distractors | Ratio of Invalid Distractors |
|---|---|---|
| Confidence ranking w/o validity rules | 482 | 53.2% |
| Confidence-ranking Control | 184 | 20.3% |
| 3-Factor Ranking Control | 160 | 17.7% |

Table 7. Manual annotation of 906 distractors generated with confidence ranking w/o validity rules, and our methods of Confidence-Ranking Control and 3-Factor Ranking Control

The following are examples of items with the answer (bolded) and invalid distractors (italicized) generated by confidence ranking without validity rules. The same item with distractors generated using Confidence-ranking Control and 3-Factor Ranking Control is also shown below:

**Example #1:**

I hope I did the right thing, Mom, Alice said. I saw a cat, all bloody but alive. I [MASK] it to the vet's, and was asked to make payment immediately.

(1) Original options:
   A. **carried**  B. followed  C. returned  D. guided

(2) Distractors generated without control:
   A. **carried**  B. *took*  C. brought  D. delivered

(3) Distractors generated with 3-Factor Ranking Control:
   A. **carried**  B. showed  C. reported  D. tried

(4) Distractors generated with Confidence Ranking Control:
   A. **carried**  B. transported  C. hauled  D. rode

**Example #2:**

[MASK] this surprised him very much, he went through the paper twice, but was still not able to find more than one mistake, so he sent for the student to question him about his work after the exam.

(1) Original options:
   A. **As**  B. For  C. So  D. Though

(2) Distractors generated without control:
   A. **As**  B. *Because*  C. Although  D. Though

(3) Distractors generated with 3-Factor Ranking Control:
   A. **As**  B. Even  C. Once  D. Soon

(4) Distractors generated with Confidence Ranking Control:
   A. **As**  B. Realizing  C. Again  D. Initially

## C  Instruction to Annotators for Invalid Distractor Identification.

**Instruction**: You are given a set of multiple-choice cloze test questions, each with four options. The correct answer is identified, along with three generated distractor options. Please review the choices and identify any "invalid distractors" - alternatives that contextually fit the gap as a potentially correct response, rather than an implausible one.

For example:

--------------------------

When I began planning to move to Auckland to study, my mother was worried about a lack of jobs and cultural differences. Ignoring these ____ I got there in July 2010.

A. **concerns**       B. *worries*

C. fears            D. considerations

--------------------------

Here, the answer is "concerns". The generated distractors include "worries". Both are grammatically correct. "Concerns" fits the semantic context only slightly better. Therefore, in this case, "worries" is considered an "invalid distractor".

Your annotation results will help assess the efficacy of our difficulty-control strategies in limiting invalid distractor generation for multiple choice cloze tests.